# Using Deep Convolutional Networks for Gesture Recognition in American Sign Language

**Vivek Bheda** and **N. Dianna Radpour**

Department of Computer Science, Department of Linguistics

State University of New York at Buffalo

{vivekkan, diannara}@buffalo.edu

## Abstract

In the realm of multimodal communication, sign language is, and continues to be, one of the most understudied areas. In line with recent advances in the field of deep learning, there are far reaching implications and applications that neural networks can have for sign language interpretation. In this paper, we present a method for using deep convolutional networks to classify images of both the the letters and digits in American Sign Language.

## 1. Introduction

Sign Language is a unique type of communication that often goes understudied. While the translation process between signs and a spoken or written language is formally called 'interpretation,' the function that interpreting plays is the same as that of translation for a spoken language. In our research, we look at American Sign Language (ASL), which is used in the USA and in English-speaking Canada and has many different dialects. There are 22 handshapes that correspond to the 26 letters of the alphabet, and you can sign the 10 digits on one hand.

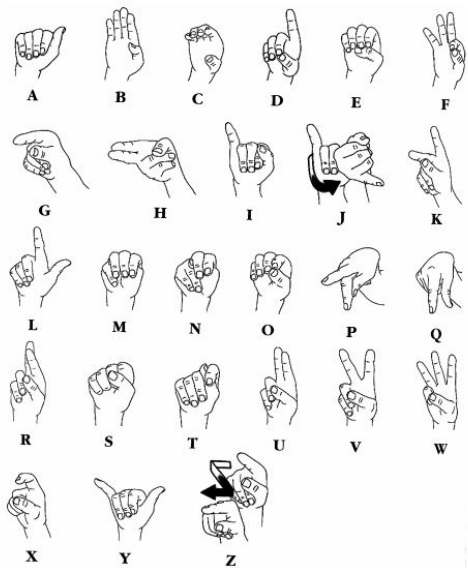

*Figure 1. American Sign Language Alphabet*

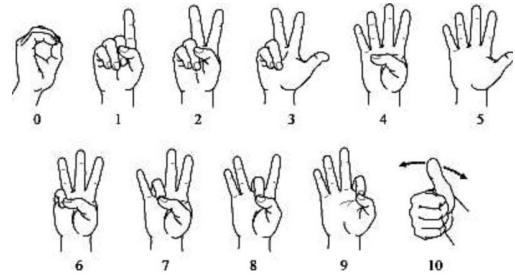

*Figure 2. American Sign Language Numbers*

One of the nuances in sign language is how often fingerspelling is used. Fingerspelling is a method of spelling words using only hand gestures. One of the reasons the fingerspelling alphabet plays such a vital role in sign language is that signers used it to spell out names of anything for which there is not a sign. People's names, places, titles, brands, new foods, and uncommon animals or plants all fall broadly under this category, and this list is by no means exhaustive. Due to this reason, the recognition process for each individual letter plays quite a crucial role in its interpretation.

## 2. Related Work

Convolutional Neural Networks have been extremely successful in image recognition and classification problems, and have been successfully implemented for human gesture recognition in recent years. In particular, there has been work done in the realm of sign language recognition using deep CNNs, with input-recognition that is sensitive to more than just pixels of the images. With the use of cameras that sense depth and contour, the process is made much easier via developing characteristic depth and motion profiles for each sign language gesture [5].

The use of depth-sensing technology is quickly growing in popularity, and other tools have been incorporated into the process that have proven successful. Developments such as custom-designed color gloves have been used to facilitate the recognition process and make the feature extraction step more efficient by making certain gestural units easier to identify and classify [8].

Until recently, however, methods of automatic sign language recognition weren't able to make use of the depth-sensing technology that is as widely available today. Previous works made use of very basic camera technology to generate datasets of simply images, with no depth or contour information available, just the pixels present. Attempts at using CNNs to handle the task of classifying images of ASL letter gestures have had some success [7], but using a pre-trained GoogLeNet architecture.

## 3. Method

Our overarching approach was one of basic supervised learning using mini-batch stochastic gradient descent. Our task was that of classification using deep convolutional neural networks to classify every letter and the digits, 0-9, in ASL. The inputs were fixed size high-pixel images, 200 by 200 or 400 by 400, being padded and resized to 200 by 200.

### 3.1 Architecture

Most implementations surrounding this task have attempted it via transfer learning, but our network was trained from scratch. Our general architecture was a fairly common CNN architecture, consisting of multiple convolutional and dense layers. The architecture included 3 groups of 2 convolutional layers followed by a max-pool layer and a dropout layer, and two groups of fully connected layer followed by a dropout layer and one final output layer.

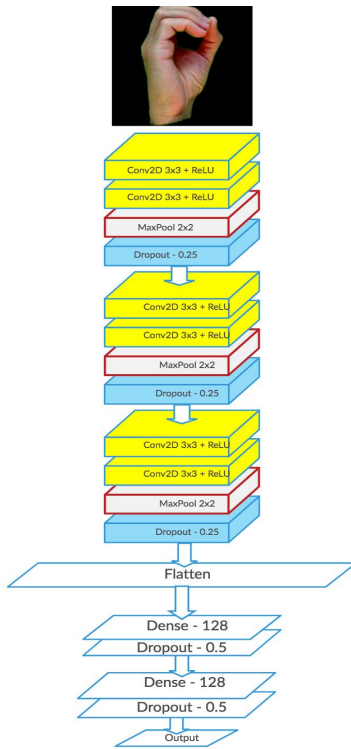

*Figure 3. Network Architecture*

## 4. Data

We initially trained and tested on a self-generated dataset of images we took ourselves. This dataset was a collection of 25 images from 5 people for each alphabet and the digits 1-9. Since our dataset was not constructed in a controlled setting, it was especially prone to differences in light, skin color, and other differences in the environment that the images were captured in, so we also used a premade dataset to compare our dataset's performance with [3]. Additionally, a pipeline was developed that can be used so people are able to generate and continue adding images to this dataset.

### 4.1 Preprocessing

For generating our own dataset, we captured the images for each sign, then removed the backgrounds from each of the images using background-subtraction techniques. When we initially split the dataset into two for training and validation, the validation accuracy showed to be high. However, when we used datasets from two different sources, i.e. training on ours and testing on the premade and vice versa, the validation accuracy drastically decreased. Since training on one dataset and validating on another was not yielding as accurate of results, we used the premade dataset for the different gestures to train the network which yielded the following results.

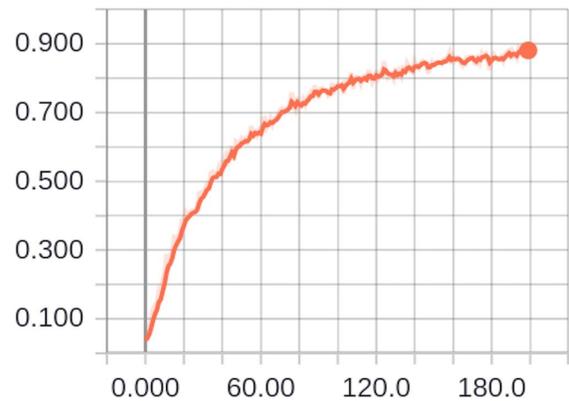

*Figure 4. Training Accuracy on ASL Alphabets*

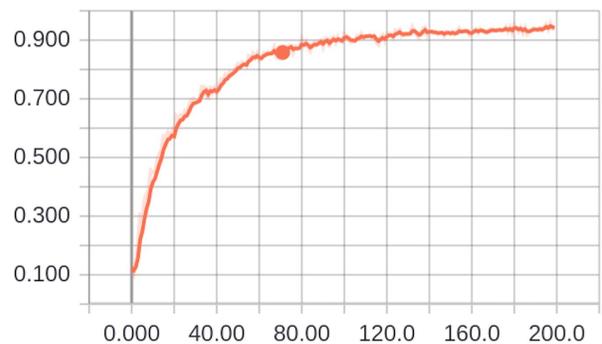

*Figure 5. Training Accuracy on ASL Digits*

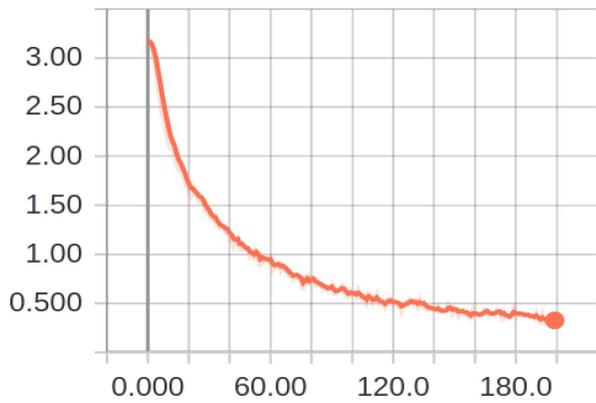

*Figure 6. Training Loss on ASL Alphabet*

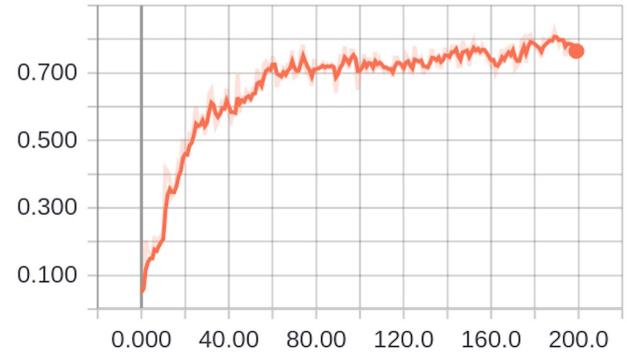

*Figure 8. Validation Accuracy on ASL Letters*

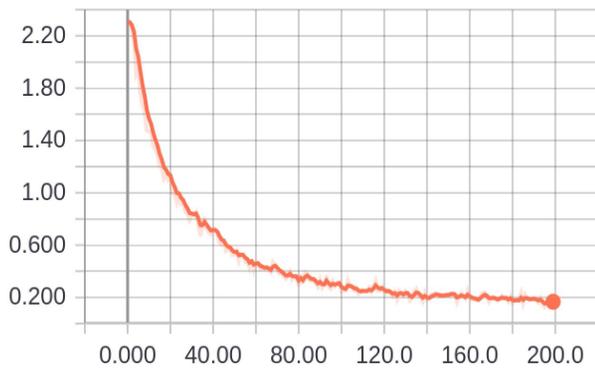

*Figure 7. Training Loss on ASL Digits*

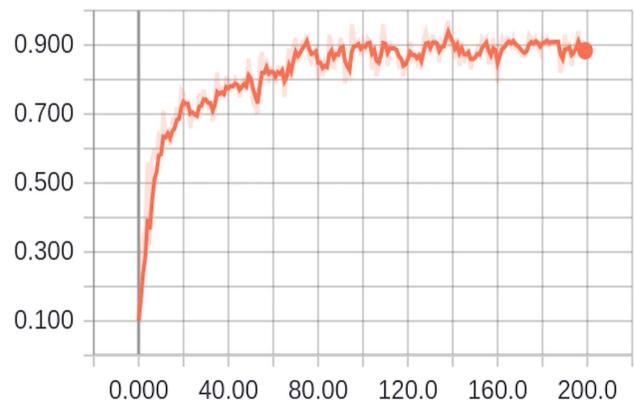

*Figure 9. Validation Accuracy on ASL Digits*

### 4.2 Data Augmentation

We saw the performances improve differently in our two datasets via data augmentation. By transforming our images just a few pixels (rotating by 20 degrees, translating by 20% on both axes) there was an increased accuracy of approximately 0.05. We also flipped the images horizontally as we can sign using both hands. While it wasn't extremely effective, we saw that with better and more representative initial training data, augmenting improved the performance more drastically. This was observed after augmentation of the premade dataset, which improved the performance by nearly 20%.

### 5. Results

We observed 82.5% accuracy on the alphabet gestures, and 97% validation set accuracy on digits, when using the NZ ASL dataset. On our self-generated dataset, we observed much lower accuracy measures, as was expected since our data was less uniform than that which was collected under studio settings with better equipment. We saw 67% accuracy on letters of the alphabet, and 70% accuracy on the digits. In terms of time complexity, gestures of the letters converged in approximately 25 minutes, and the digits converged in nearly 10 minutes.

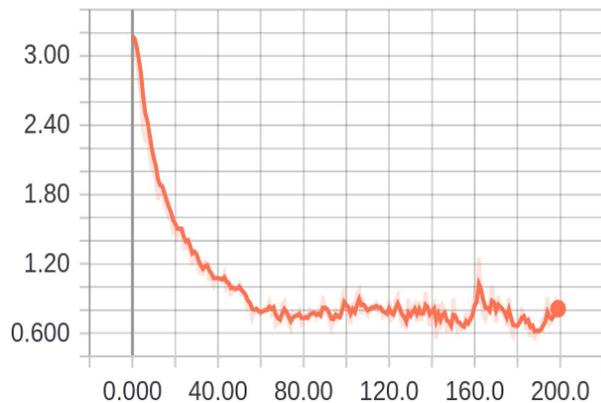

*Figure 10. Validation Loss on ASL Letters*

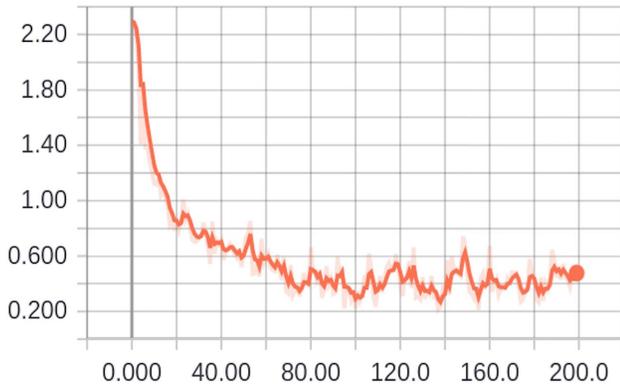
*Figure 11. Validation Loss on ASL Digits*

**5.1 Evaluation**
We trained with a categorical cross entropy loss function for both our datasets. It is a fairly common loss function used along with image classification problems.

$$H(p,q) = -\sum_{x} p(x)\log(q(x))$$

Initially, we observed low accuracy measures when testing on the validation set of the self-generated data, which we accounted largely to the lighting and skin tone variations in the images. The higher accuracy measure for the digits was expected, since the gestures for the digits are much more distinguishable and easier to classify. Compared to previous methods working on this same task, our network performed quite well, considering RF-JA were using both a color glove and depth-sensing Kinect camera. The cause of higher accuracy than Stanford's method was likely due to their lack of background-subtraction for the images, since they used a large dataset from ILSVRC2012 as part of a competition.

| Method | Accuracy |
|---|---|
| deepCNN (our method) | 82.5 |
| Stanford deepCNN [7] | 72 |
| RF-JA+C(h-h) [8] | 90 |
| RF-JA+C(l-o-o) [8] | 70 |

*Figure 12. Comparison of previous methods with ours; Stanford didn't use background subtraction, RF-JA(h-h) split the training and validation set 50-50, (l-o-o) omitted specific data.*

**6. Conclusions and Future Work**
In this paper, we described a deep learning approach for a classification algorithm of American Sign Language. Our results and process were severely affected and hindered by skin color and lighting variations in our self-generated data which led us to resort to a pre-made professionally constructed dataset. With a camera like Microsoft's Kinect that has a depth sensor, this problem is easy to solve [5]. However, such cameras and technology are not widely accessible, and can be costly. Our method shows to have potential in solving this problem using a simple camera, if enough substantial training data is provided, which can be continuously done and added via the aforementioned processing pipeline. Since more people have access to simple camera technologies, this could contribute to a scalable solution.

In recognizing that classification is a limited goal, we plan on incorporating structured PGMs in future implementations of this classification schema that would describe the probability distributions of the different letters' occurrences based on their sequential contexts. We think that by accounting for how the individual letters interact with each other directly (e.g. the likelihood for the vowel 'O' to proceed the letter 'J'), the accuracy of the classification would increase. This HMM approach with sequential pattern boosting (SP-boosting) has been done with the actual gesture units that occur in certain gestures' contexts, i.e. capturing the upper-arm movements that precede a certain letter to incorporate that probability weight into the next unit's class, [6] and processing sequential phonological information in tandem with gesture recognition [4], but not for part-of-word tagging with an application like what we hope to achieve.

We also recognize that the representation itself makes a huge difference in the performance of algorithms like ours, so we hope to find the best representation of our data, and building off our results from this research, incorporate it into a zero-shot learning process. We see zero-shot learning as having the potential to facilitate the translation process from American Sign Language into English. Implementing one-shot learning for translating the alphabet and numbers from American Sign Language to written English, and comparing it with a pure deep learning heuristic could be successful and have the potential to benefit from error correction via language models. Recent implementations of one-shot adaptation have also had success in solving real world computer vision tasks, and effectively trained deep convolutional neural networks using very little domain-specific data, even as limited as single-image datasets. We ultimately aim to create a holistic and comprehensive representation learning system for which we have designed a set of features that can be recognized from simple gesture images that will optimize the translation process.

## 7. References

[1] X. Chen and A. Yuille. Articulated pose estimation by a graphical model with image dependent pairwise relations. In *Advances in Neural Information Processing Systems (NIPS)*, 2014.

[2] T. Pfister, J. Charles, and A. Zisserman. Flowing convnets for human pose estimation in videos. In *IEEE International Conference on Computer Vision*, 2015.

[3] Barczak, A.L.C., Reyes, N.H., Abastillas, M., Piccio, A., Susnjak, T. (2011), A new 2D static hand gesture colour image dataset for ASL gestures, *Research Letters in the Information and Mathematical Sciences*, 15, 12-20

[4] Kim, Taehwan & Livescu, K & Shakhnarovich, Greg. (2012). American sign language fingerspelling recognition with phonological feature-based tandem models. In *IEEE Spoken Language Technology Workshop (SLT)*, 119-124.

[5] Agarwal, Anant & Thakur, Manish. Sign Language Recognition using Microsoft Kinect. In *IEEE International Conference on Contemporary Computing*, 2013.

[6] Cooper, H., Ong, E.J., Pugeault, N., Bowden, R.: Sign language recognition using sub-units. *The Journal of Machine Learning Research*, 13(1), 2205–2231, 2012.

[7] Garcia, Brandon and Viesca, Sigberto. Real-time American Sign Language Recognition with Convolutional Neural Networks. In *Convolutional Neural Networks for Visual Recognition* at Stanford University, 2016.

[8] Cao Dong, Ming C. Leu and Zhaozheng Yin. American Sign Language Alphabet Recognition Using Microsoft Kinect. In *IEEE International Conference on Computer Vision and Pattern Recognition Workshops*, 2015.